\documentclass[11pt,a4paper]{article}

\usepackage[pdfpagelabels=false]{hyperref}
\usepackage[hyperref]{acl2019}

\usepackage{times}
\usepackage{latexsym}

\usepackage{soul}
\usepackage{bbm}

\usepackage{amsmath}
\usepackage{amssymb}
\usepackage{bm}
\usepackage{makeidx}
\usepackage{graphicx}
\usepackage{siunitx}
\usepackage{color}
\usepackage{colortbl}
\usepackage{textcomp}
\usepackage{gensymb}
\usepackage{breqn}

\usepackage{diagbox}
\usepackage{subcaption}
\usepackage{listings}

\usepackage[]{algorithm2e}

\usepackage{url}

\usepackage{array}
\newcolumntype{P}[1]{>{\centering\arraybackslash}p{#1}}
\newcolumntype{M}[1]{>{\centering\arraybackslash}m{#1}}
\newcolumntype{N}{@{}m{0pt}@{}}
\usepackage{multirow}

\usepackage{float}
\aclfinalcopy 

\title{EQuANt (Enhanced Question Answer Network)}

\author{Fran\c cois-Xavier Aubet \thanks{\ \  All authors contributed equally to this work. Correspondence to francois-xavier.aubet.18@ucl.ac.uk} \\
  {University College London}     \And
  Dominic Danks \footnotemark[1] \\
  {University College London}
  \And
  Yuchen Zhu \footnotemark[1] \\
  {University College London} \\}

\date{}

\begin{document}
\maketitle



\begin{abstract}
Machine Reading Comprehension (MRC) is an important topic in the domain of automated question answering and in natural language processing more generally. Since the release of the SQuAD 1.1 and SQuAD 2 datasets, progress in the field has been particularly significant, with current state-of-the-art models now exhibiting near-human performance at both answering well-posed questions and detecting questions which are unanswerable given a corresponding context. In this work, we present \textbf{E}nhanced \textbf{Qu}estion \textbf{A}nswer \textbf{N}e\textbf{t}work (EQuANt), an MRC model which extends the successful QANet architecture of \cite{qanet} to cope with unanswerable questions. By training and evaluating EQuANt on SQuAD 2, we show that it is indeed possible to extend QANet to the unanswerable domain. We achieve results which are close to $2 \times$ better than our chosen baseline obtained by evaluating a lightweight version of the original QANet architecture on SQuAD 2. In addition, we report that the performance of EQuANt on SQuAD 1.1 after being trained on SQuAD2 exceeds that of our lightweight QANet architecture trained and evaluated on SQuAD 1.1, demonstrating the utility of multi-task learning in the MRC context.
\end{abstract}

\section{Introduction} 
\label{sec:introduction}

Machine Reading Comprehension (MRC) entails engineering an agent to answer a query about a given context. The complexity of the task comes from the need for the agent to understand both the question and the context. Progress has been largely driven by datasets that have addressed increasingly difficult intermediate tasks. In particular, the SQuAD 1.1 dataset \cite{squad1} was released in 2016, providing an extensive set of paragraphs, questions and answers. As models rivalled human performance on that dataset, SQuAD 2 was released with an additional 50,000 adversarially written unanswerable questions. 

Motivated by the general question of how an MRC agent can be adapted when its original MRC task assumptions are relaxed, we work on the specific research problem of relaxing the answerability assumption on the MRC task, and we evaluate our work using the SQuAD 2 dataset.

QANet \cite{qanet} is a feedforward architecture using only convolutions and attention mechanisms for MRC. It is devoid of recurrence, which is a typical ingredient in previous MRC models, and despite its simplicity it achieved state-of-the-art performance on SQuAD 1.1. Observing the absence of a mechanism in QANet to allow for unanswerability, and noting that to the best of our knowledge there has so far been no effort to incorporate one, we decided to base our work on this architecture. Our contribution is two-fold:


Firstly, we present EQuANt, which extends the original QANet architecture to include an answerability module. Working within the time and resource constraints of this project, we achieved a 63.5 F1 score on SQuAD 2, almost double the accuracy of our baseline QANet method. For the sake of reproducibility, we make available an open-source implementation of our model at \url{https://github.com/Francois-Aubet/EQuANt}.



Secondly, we show that by training EQuANt to accomplish two distinct tasks simultaneously, namely answerability prediction and answer extraction, we improve the model's performance on SQuAD 1.1 from that of QANet, verifying that a multitask learning approach can improve an MRC model's performance.

We begin in section \ref{sec:Background} by presenting the background necessary to motivate and understand our contribution. In section \ref{sec:Related Work}, we give an overview of related work and how it complements and differs from our work. In sections \ref{sec:Methods}, \ref{sec: experiment} and \ref{sec: results} we illustrate the design of our model and present and discuss our experimental results. Finally, in section \ref{sec:Conclusion}, we summarise our work and propose potential future work which would extend our contribution.

\section{Background}
\label{sec:Background}

The problem of question answering can be formulated specifically in the open domain setting in the following way:

\textit{Given a question, or query, sequence $Q = (q_1, ..., q_{m})$, and a context paragraph sequence $C = (c_1, ...,c_{n})$, assume the answer to the question is a unique connected subsequence of $C$, then identify that subsequence. i.e. Identify $i,j\in \{1,..., n\}$, $i \leq j$, such that the span $A = (c_{i},..., c_{j})$ is the answer to the query $Q$.} \hspace*{\fill}  ($\star$)

A recent and significant dataset responsible for much of the development of models in tackling the above-formulated problem is the Stanford Question Answering Dataset (SQuAD), and more specifically its two versions, SQuAD 1.0 and SQuAD 1.1, \cite{squad1}. SQuAD consists of over 100,000 crowdsourced comprehension questions and answers based on Wikipedia articles. Importantly, this dataset is large enough to support complex deep learning models and contains a mixture of long- and short-phrase answers which are directly implied by the associated passage. Since its introduction, SQuAD has inspired healthy competition among researchers to hold the state-of-the-art position on its leaderboard.

The success of an MRC model hinges on its ability to represent both the structures of the questions and contexts, and the relationship between the questions and the contexts. The two most prominent methods in the literature to represent the structures of such kinds of sequential data are attention and recurrence, thus it is not surprising that the best performing models on SQuAD 1.0 leaderboard are attention-based models,
e.g. BERT \cite{BERT}, and RNN-based models, e.g. R\-Net, \cite{rnet}. One prominent attention-based candidate on the leaderboard is QANet, \cite{qanet}, upon which our work is built. We will now provide a brief introduction to QANet and motivate our decision to work with this model.

\begin{figure}[t]
    \centering
    \includegraphics[scale=0.6]{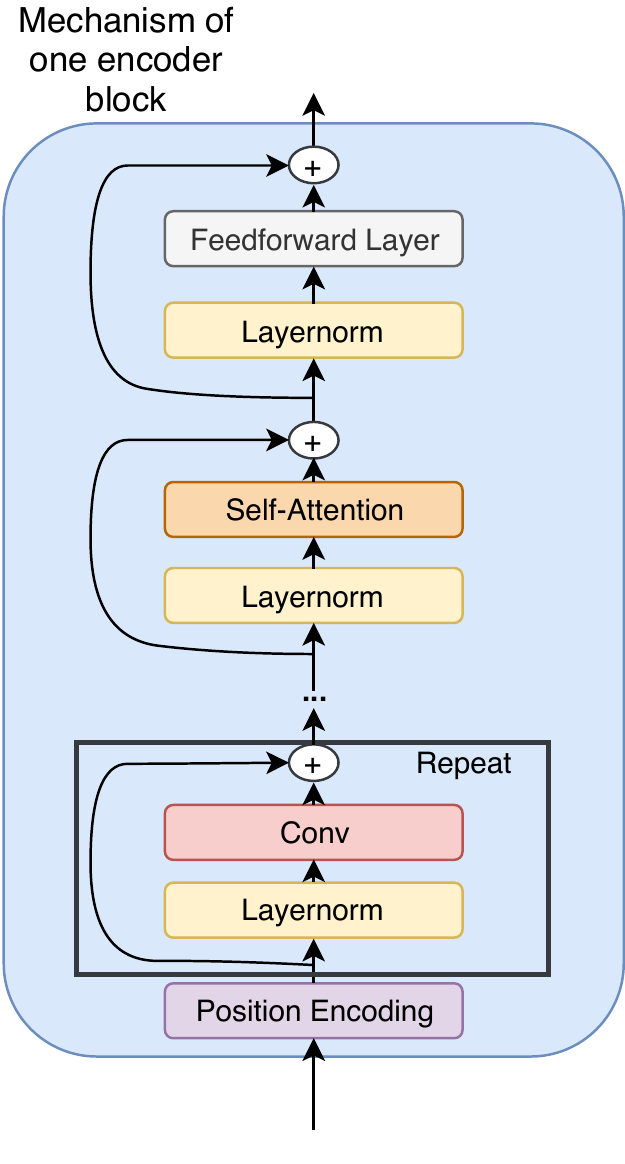}
    \caption{The mechanism of one QANet encoder block.}
    \label{fig:1_encoder}
\end{figure}

QANet consists of five functional blocks: a context processing block, a question processing block, a context-query block, a start-probability block and a end-probability block. See figure \ref{fig:equant} for a high level representation of the model. Within the context, question and context-query blocks, an embedding encoder of the form shown in figure \ref{fig:1_encoder} is used repeatedly. This is very similar to the transformer encoder block introduced in \cite{need_paper}, however possesses an additional convolutional layer after positional encoding and before the layernorm and self-attention layer.
These additional separable convolutional layers enable the model to capture local structure of the input data. Having passed through the context-query block, the data is then passed into the two probability blocks, which are both standard feed-forward layers with softmax, to calculate the probability of each word being a start- or end-word. For a detailed description of each portion of the model, the reader is referred to the original paper \cite{qanet}, however further discussion of the components most relevant to our architecture design and experiments can be found in section \ref{subsec:methods_base_qa}.

The original QANet authors \cite{qanet} achieved a result of 73.6 Exact Match (EM)
and 82.7 F1 score on the SQuAD 1 datasets, placing it among the then state-of-the-art models. This is quite remarkable given its apparent conceptual and practical simplicity. Armed with separable convolution and an absence of recurrence, it is able to achieve its results whilst having a faster training and inference time than all of the RNN-based models preceding it \cite{qanet}. Thus we are motivated to investigate the properties of a simple, efficient and accurate model in hope of gaining fundamental understanding of question-answer modelling.

As methods on the top of the SQuAD 1.1 leaderboard started to outperform human EM and F1 scores, a more challenging task was called for, leading to the entrance of SQuAD 2.0 \cite{squad2}. In addition to SQuAD 1.1, SQuAD 2.0 added over 50,000 unanswerable questions written adversarially to look similar to answerable ones. 

Under this new setting, we reformulate the question answering problem as the following:

\noindent \textit{For $Q$ and $C$ as given in $(\star)$, release the assumption that an answer exists, but if it does then assume it is a unique connected subsequence of $C$.
First identify the value of the indicator variable $b \in \{0, 1\}$, such that if the answer exists, then $b=1$, otherwise $b=0$. Furthermore, if the context contains the answer, then identify $i,j \in \{1,...,n\}$, $i \leq j$, such that the span $A = {c_i,...,c_j}$ is the answer to the query $Q$.} \hspace*{\fill}($\star \star$)

Inspecting the QANet architecture, it is not hard to see that the model would not give the desired prediction for unanswerable questions, as the value of $b$ is assumed to be $1$ and the length of the span is at least $1$. This motivates our extension of QANet to accommodate for unanswerable questions and identifies SQuAD 2.0 as an appropriate benchmark dataset for our work.



\section{Related Work}
\label{sec:Related Work}
\subsection{Open domain question answering}

 Recurrence has traditionally been a key component in many natural language tasks, including QA, and since the entrance of attention mechanisms into the QA domain, models have also found success by using attention to guide recurrence, such as the 
 BiDAF and SAN models \cite{BiDAF,SAN}. A key drawback in traditional recurrence-based architectures is the long training time due to the $O(n)$ complexity in modelling relations between words which are $n$ words apart. Replacing recurrence with pure attention completely reduces the complexity to constant, providing faster algorithms \cite{need_paper}.

SAN, alongside other models, e.g. \cite{hill16, dhingral16}, used multi-step reasoning, implemented using recurrent layers, to predict the answer spans. The purpose of using multiple reasoning states is to extract higher order logical relations in contexts. For example, the model may first learn what a pronoun is referring to before extracting the answer span based upon the reference. In contrast, QANet used entirely multi-headed attention and convolution mechanisms, which encapsulate the intra-context relations, and is in addition superior at modelling long-range relations. Moreover, the large recurrence component of these models creates a burden on training speed, whereas QANet's attention and separable convolution approach saves an order of magnitude on the training complexity by the result stated in the first paragraph of this section. 



The transformer architecture proposed in
\cite{need_paper} and its 
pre-trained counterpart, BERT \cite{BERT}, have become the common factor in all leading QA models. Unlike QANet, which is specifically designed for QA, BERT is an all-in-one model, capable of aiding many natural language tasks. Thus it is not surprising that BERT is a much larger model than QANet, containing $110M$ parameters in the base model, compared with the fewer than $5M$ parameters that were present in QANet during our computational experiments. As a result, BERT will have significantly greater inference time than QANet. Furthermore, as a result of its multi-faceted abilities, impressive though they are, BERT is less capable of illustrating the interaction of the model with the QA problem specifically. QANet, however, as a simple feed-forward, QA-specific model with less training time, allows more insights into the model's reaction to the problem, hence providing researchers with more intuitions into model enhancement.

\subsection{Unanswerability extension}

One body of unanswerability extension relies on incorporating a no-answer score to the model, which is the main inspiration for our work. Levy et al. extended BiDAF \cite{Levy_et_al, BiDAF} to deal with unanswerable questions by effectively setting a threshold $p$, whereby the model will output no answer if the model's highest confidence in any answers is less than $p$. Our work uses an approach similar to Levy's to verify that QANet generates generally lower probabilities in a dataset with unanswerable questions, but our final model adopts a more explicit approach.

In \cite{extended_SAN}, the original SAN authors extended SAN to accommodate unanswerable questions. Their work added an extra feed-forward module for discrimination of answerable/unanswerable questions and trained an objective which jointly accounts for answer correctness and answerability. We take inspiration from this extended SAN, but the summary statistic fed into the answerability module in our model is obtained from a fundamentally different procedure which is completely devoid of recurrence. We also favour the approach of minimising a joint objective over different tasks in response to recent successes of multi-task approaches in NLP which suggest that a learner's learning generalisability improves as it tries to accomplish more than one task.


Read+Verify and UNet \cite{read_verify, unet} both use an additional answer verifier to improve performance by finding local entailment that supports the answer by comparing the answer sentence with the question. The local entailment finding is able to improve the answer accuracy as it is sensitive to specific types of unanswerability, such as an impossible condition. Due to time constraints, we leave the exploration of utilising verification modules for future work.


\section{Methods}
\label{sec:Methods}

\subsection{Light QANet implementation}\label{subsec:methods_base_qa}
Given that our model builds directly upon QANet, a natural first step
was to work with a computational implementation of this base model. We chose to
use the open-source Tensorflow implementation of QANet hosted at \url{https://github.com/NLPLearn/QANet}
for our QANet experiments and as a base for our extension. A particularly attractive
aspect of this implementation is that it allows straightforward customisation of the
hyperparameters involved in the QANet model. In order to allow for a larger number of
design iterations and to account for limited computational resources, we chose to utilise
this customisability to make our trained QANet model lightweight, and we refer to this
scaled-down version of the original QANet as ``light QANet". Although this choice is likely
to mean that our results, quoted in section \ref{sec: results}, could be improved by increasing
the complexity of the architecture, we stress that our aim is not to surpass state-of-the-art
performance on SQuAD2, but instead to show that it is possible to successfully extend QANet
to the unanswerable domain.

\begin{figure}[t]
    \centering
    \includegraphics[scale = 0.35]{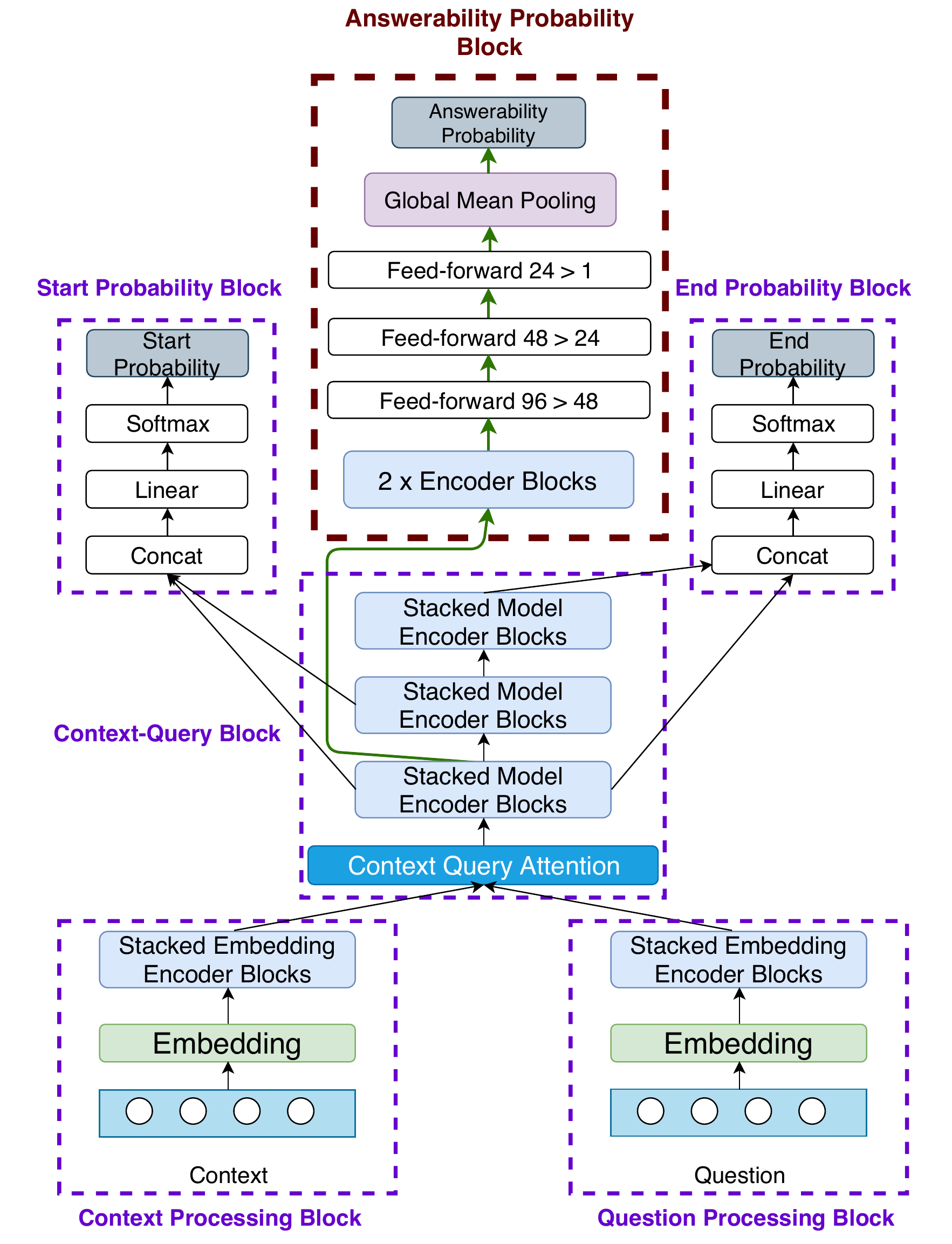}
    \caption{EQuANt architecture: combination of QANet and unanswerability extension module.}
    \label{fig:equant}
\end{figure}

As in the original paper, our character embeddings are trainable and are initialised by
truncating GloVe vectors \cite{glove}. However, in the interest of model size, we choose to retain
$p_2' = 64$ of the $p_1 = 300$ of each GloVe vector rather than $p_2 = 200$ as
in the original paper. We then used the character embedding convolution of QANet to map the $p_2' = 64$-dimensional vector to a $p_2'' = 96$-dimensional
representation, making the output of
our context and query input embedding layers of dimension
$p_1 + p_2'' = 396$, rather than $p_1 + p_2 = 500$ used by the original authors, resulting
in a significant reduction in the number of parameters in our model. Having utilised the input embedding to represent each word in the context and query
as a $396$-dimensional vector, these vectors then flow into the embedding encoder blocks
(of the form shown in \ref{fig:1_encoder}),
where they are transformed using a series of convolutions, feedforward layers and self-attention.
In these encoder blocks and throughout the rest of the network, we choose our hidden layer size
to be $96$, as opposed to original QANet's hidden layer size of $128$. Furthermore,
although the typical transformer architecture relies on multi-headed self-attention,
with the original QANet using $8$ heads in all layers,
this introduces additional computational overhead. As a result, we minimise this
by using only a single head, however it is straightforward to change the number of heads in our implementation and this may yield fruitful results. All other architecture and hyperparameter choices match those described in the original paper \cite{qanet}.


As well as using this process to gain an understanding of the inner workings of QANet,
we utilised light QANet to provide a principled initialisation for the training of our
extended architecture. In \cite{qanet}, the authors describe how they used an augmented
dataset generated using neural machine translation and how this significantly improved their
results. As having access to this dataset would likely result in improved outcomes for our
model, we initiated contact with the QANet authors, however access to the augmented dataset
was not granted. As a result, we trained light QANet on SQuAD 1.1 for $32,000$ iterations, providing
the results shown in table \ref{table:squad2results} and saved the corresponding weights, allowing them to be restored and used as principled initialisations when performing subsequent model training.


\subsection{Problem Analysis}
\label{subsec: prob_analysis}

In order to gain a better understanding of our research problem both conceptually and practically, and to assess our intuition to extend QANet with an extra answerability module, we specifically ask ourselves two sub-questions: 1) How much does QANet detect distinction between answerable and unanswerable questions? 2) What should be chosen as the input to our answerability module?

We first ran our light QANet on answerable and unanswerable questions extracted from the SQuAD 2 dataset 
and analysed the results. 

\begin{figure}[t]
    \centering
    \includegraphics[scale = 0.35]{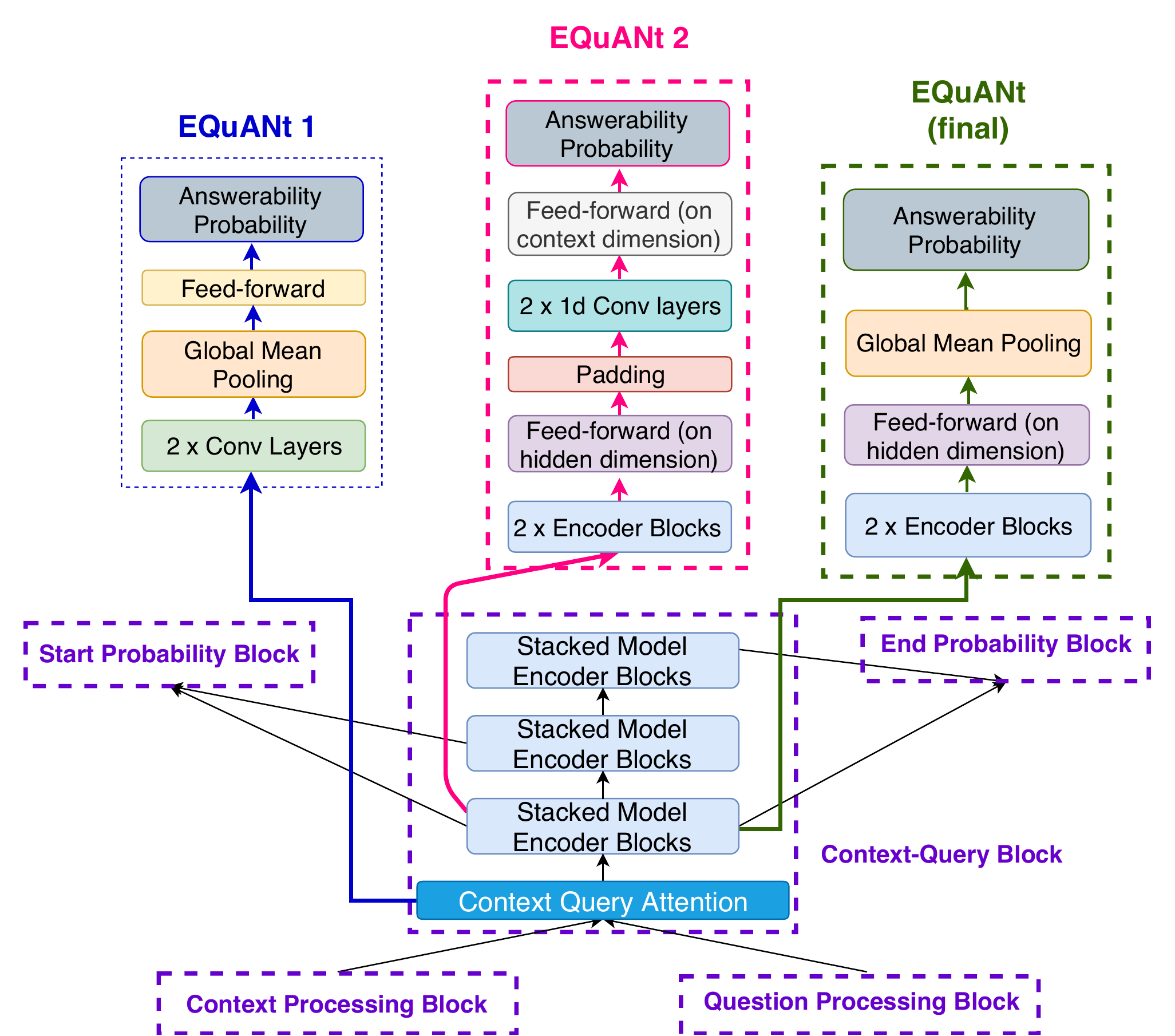}
    \caption{Attempts to extend QANet to EQuANt.}
    \label{fig:attempts}
\end{figure}


Our results showed that QANet assigns generally lower probability to all possible ``answers" to unanswerable questions. More precisely, given context-query pairs from answerable and unanswerable questions, we study the maximum start-word and end-word probabilities assigned by QANet to all words in the context, and we find that unanswerable questions on average receive lower start- and end-word probabilities on all words in the corresponding context. This shows that the original QANet already captures information about unanswerability, validating the possibility of answerability detection by appending an additional functional module to the basic QANet structure.

Upon inspection of the intermediate outputs of the QANet architecture, we found that QANet respects the variable length of input queries and contexts, resulting in all intermediate outputs of the architecture having variable size. Whilst this is compatible with QANet's original aim of assigning probabilities to every word in the context, it is not immediately compatible with our extension, the purpose of which is to assign an answerability score to the context as a whole. It is thus necessary to design our extension to handle variable input size. In section \ref{subsec: Architecture Design}, we outline three attempted solutions.




\subsection{Enhanced Question Answer Network (EQuANt)}
\label{subsec: Architecture Design}
We now provide details on our exact architecture design, which we name \textbf{E}nhanced \textbf{Qu}estion \textbf{A}nswer \textbf{N}e\textbf{t}work (EQuANt). EQuANt is based on light QANet, with an answerability extension module as motivated in section \ref{subsec: prob_analysis}. We investigated three extension designs, with the final one achieving promising results, in particular almost doubling light QANet's accuracy on SQuAD 2 in both EM and F1 measures. 

A component of the QANet architecture which is particularly relevant to the design
of our architectures and our analysis of the inner workings of the model is the
context-query attention layer (see figure \ref{fig:equant}). This layer takes in the encoded
context and query and combines them into a single tensor. A core aspect of this
layer is the similarity matrix, $S$, which has size $(\texttt{number of context
words} \times \texttt{number of query words})$. The $ij^{th}$ element of this matrix represents
the similarity of context word $i$ and query word $j$, calculated using the trilinear function
described in \cite{BiDAF}. This matrix is important for two reasons. Firstly, visual inspection
of its components allows interpretation of the quality of the context and question encodings.
Secondly, if the model is to be successful, $S$ must contain the
information required to determine answerability or lack thereof and represents the first point in
the network where a single tensor must contain this information, making it a natural focal point
for our architecture designs.

These architecture designs are outlined in the remainder of this section and visualised in figure \ref{fig:attempts}. Of the three designs that were implemented, the third (EQuANt 3) exhibited the best performance (see table \ref{table:squad2results}) and was therefore chosen to be our final model. Discussions regarding the performance of each architecture can be found in sections \ref{sec: experiment} and \ref{sec: results}.

\subsubsection{EQuANt 1}
EQuANt 1 takes the context-query attention weights from the context-query attention layer, which are of size \texttt{length of context $\times$ length of question}, and applies two convolutional layers followed by global mean pooling and a feedforward layer. The variable-size dimensions inherited from variable context and question lengths are reduced to 1 during global mean pooling. The final feed-forward layer then transforms the channel dimension obtained from convolution layers to size 1, giving us a scalar which we use as the score. This model performed poorly on the SQuAD 2 dev set, likely due to the information loss when convolving the context-query attention matrix. More discussion is provided in section \ref{sec: experiment}. 


\subsubsection{EQuANt 2}
The EQuANt 2 extension stems from the output of the first stacked encoder layer, making each of its inputs of dimension \texttt{length of context $\times$ number of hidden nodes}. We apply two encoder transformations as in figure \ref{fig:1_encoder} and then a feedforward network which transforms the size of the hidden layer (96) to 1, followed by padding the context-length dimension to constant length. Then we apply two layers of 1d convolution, before a final feedforward layer to map to a scalar which we take as the score. This model also performed poorly. Note that padding decreases the proportion of non-zero elements along the context-length dimension in many data points significantly, essentially causing interesting information to be compressed, potentially explaining the lack of success for this model. 


\subsubsection{EQuANt 3 (Final design)}
After learning from the failure cases of EQuANt 1 and 2, we aim to design a model which extracts more useful information from the context-query attention matrix, whilst avoiding diluting it with zeros. To extract higher level information from the context-query attention matrix, we use two more encoder blocks as in figure \ref{fig:1_encoder}, after which we down-sample the context-query dimension using global mean pooling. Our exact design is as follows: the answerability module again starts from the output of the first stacked encoder layer and two encoder transformations are also applied. The output from this then undergoes three feedforward layers, which transforms the hidden dimensions from 96 to 48, 48 to 24 and 24 to 1. Finally, a global mean pooling layer takes the variable-length dimension inherited from the context length and transforms it to size 1, giving us the answerability score. EQuANt 3 performed respectfully on the SQuAD 2 dev set, achieving 70.26\% accuracy on answerability prediction. 


\begin{figure*}[h!]
	\begin{subfigure}{1.3\columnwidth}
		\centering
		\includegraphics[angle=-90,trim=0 0 0 0,clip,width=0.99\columnwidth]{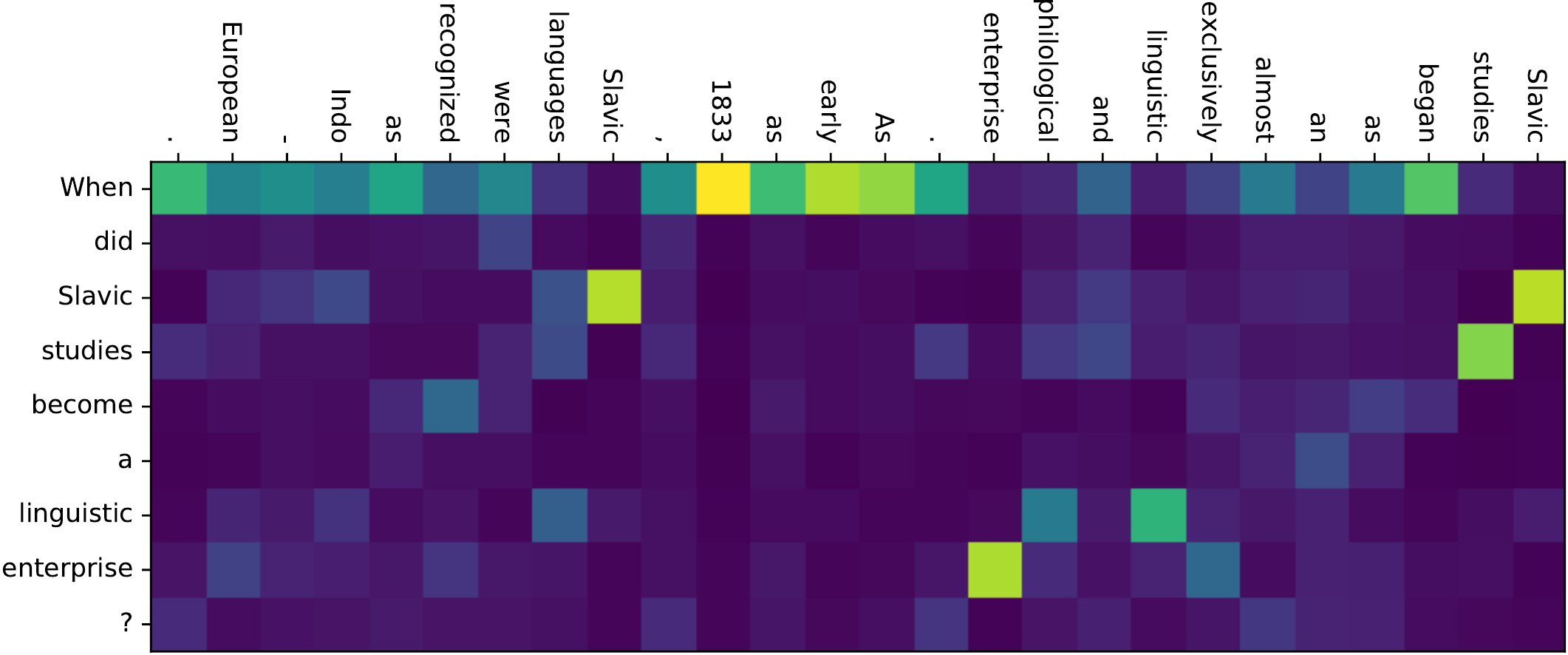}
	\end{subfigure}%
	\centering	
	
	\begin{subfigure}{1.3\columnwidth}
		\centering
		\includegraphics[angle=-90,trim=60 0 0 0,clip,width=0.99\columnwidth]{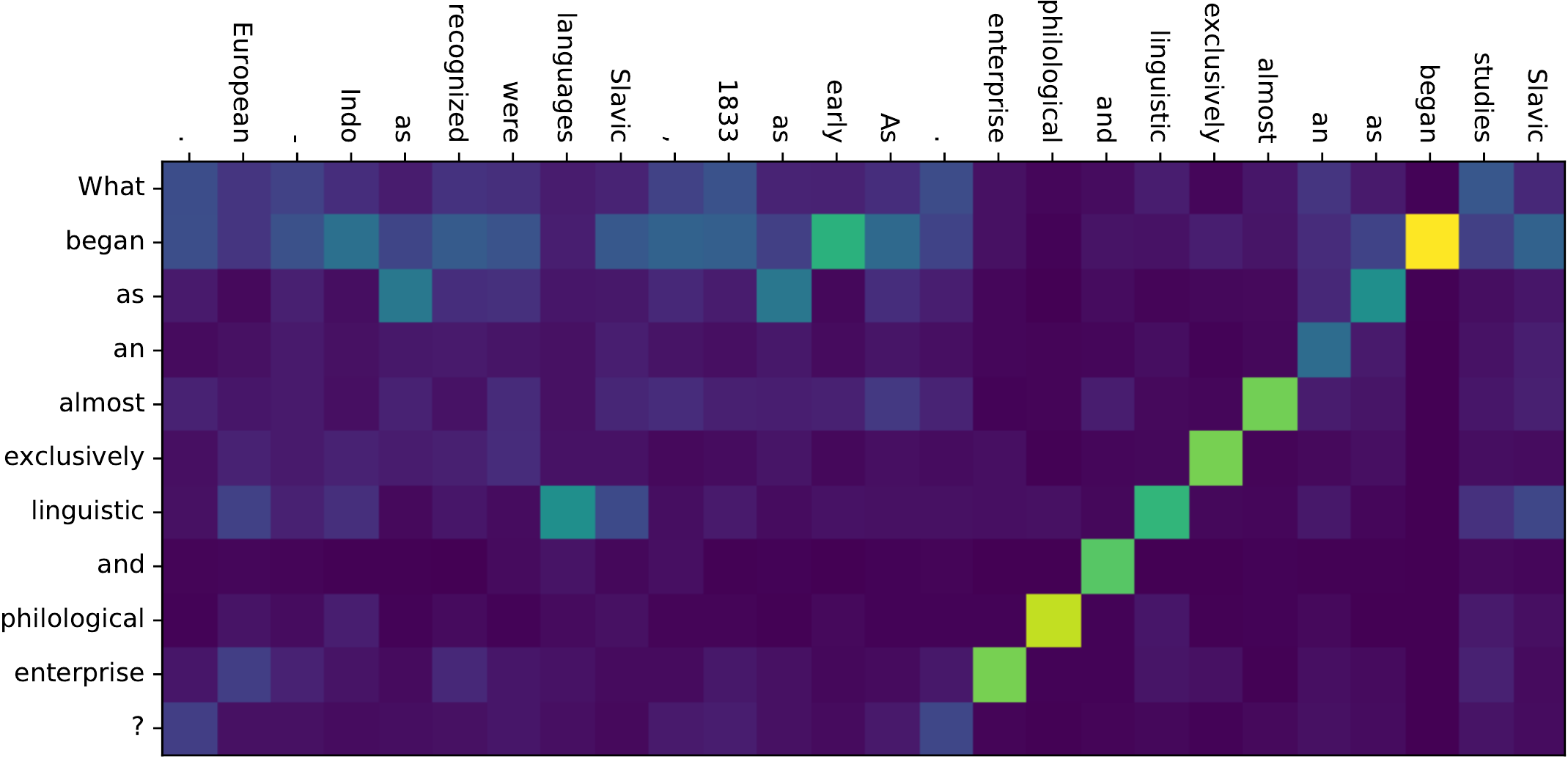}
	\end{subfigure}	
	\centering		
	
	\begin{subfigure}{1.34\columnwidth}
		\centering
		\includegraphics[angle=-90,trim=60 0 0 0,clip,width=0.99\columnwidth]{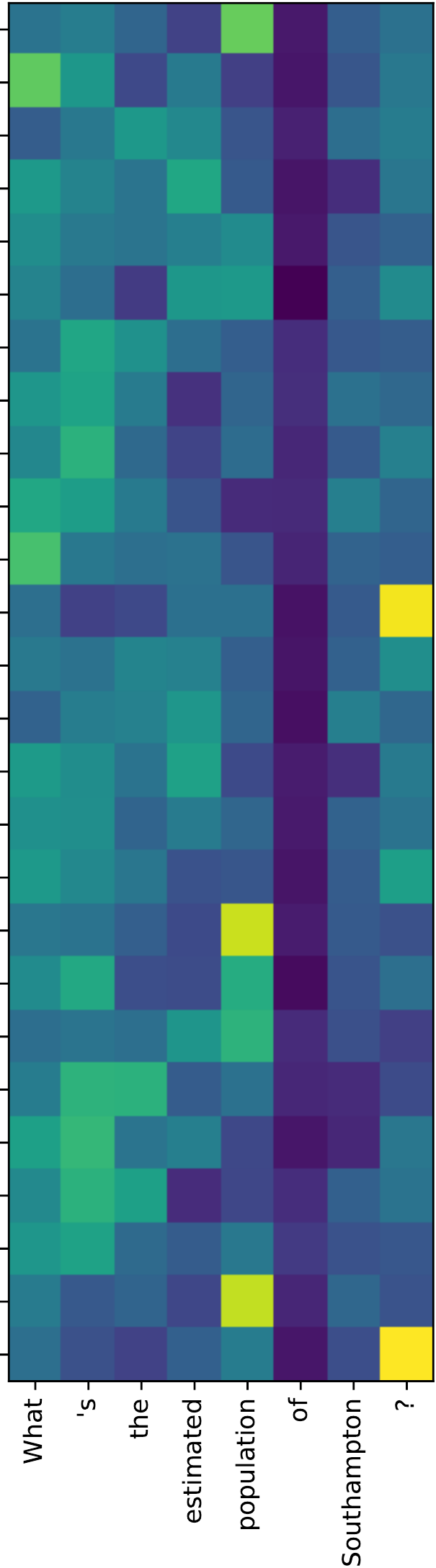}
	\end{subfigure}
	\centering
	\caption{Attention maps. Top: Unanswerable question. Middle: Answerable question. Bottom: Shuffled question.}
	\label{fig:attention_maps}
\end{figure*}

\subsubsection{Loss function}
Let $\theta$ denote the vector of parameters, $p_0$ the predicted answerability probability, $\delta$ the ground truth of answerability, $p_1$ the predicted start-word probability of the true answer and $p_2$ the predicted end-word probability of the true answer. Then our loss function can be formulated as
\begin{align*}
    l(\theta) = \frac{1}{N} \sum_{i = 1}^{N} \left[\mathcal{L}_0^{(i)}(p_0^{(i)}) + \delta^{(i)}\left(\mathcal{L}_1^{(i)}(p_1^{(i)}) + \mathcal{L}_2^{(i)}(p_2^{(i)})\right)\right],
\end{align*}
where $\mathcal{L}_j(p_j)$ with $j=1,2,3$ denotes the cross entropy loss associated with answerability, start-word and end-word predictions respectively.

In our experiments, we performed stochastic gradient descent using the Adam optimiser with hyperparameter settings: batch size = 32, learning rate = 0.001, $\epsilon = 1e-07$, $\beta_1 = 0.8$ and $\beta_2 = 0.999$.

\section{Experiments}
\label{sec: experiment}
As mentioned in section \ref{subsec: Architecture Design}, the context-query
similarity matrix, $S$, offers insight into the quality of the model's encodings and
should contain the information required to infer answerability or lack thereof.
In order to gain the most insight into potential model behaviour, we investigated
the form of $S$ within our light QANet for three different types of context-query pairs. The first two types
are the standard answerable and adversarially designed unanswerable varieties taken
directly from SQuAD 2. The final type is referred to as shuffled, for which we pair a given
context with a question from a different article, meaning that the question is almost certainly
unanswerable and unrelated to the context paragraph.

For visualisation purposes, we focused on short contexts, and an example of $S$ for a specific
short context and each of the three types of question is shown in figure \ref{fig:attention_maps}. 
These results are interesting for two main reasons. Firstly, they show that the learnt encodings
are meaningful. For example, the word ``when" in the adversarial unanswerable question attends to
the date-related part of the context, and the word ``population" in the shuffled question attends
to words in the context associated with geographical regions. Furthermore, these results perhaps
offer insights into why the initial convolution approach was unsuccessful. In particular, it seems that answerable and adversarially unanswerable questions both lead to $S$ matrices with peaked context words for each query word, making it hard for convolutions to successfully identify unanswerability. However, as expected, the $S$ matrices for shuffled questions appear more diffuse and random due to the largely unrelated meanings of the context and query, further emphasising the subtlety in distinguishing answerable and adversarially unanswerable questions.

\begin{figure*}[h!]
    \centering
	\begin{subfigure}{2\columnwidth}
	    \centering	
		\begin{tabular}{ |M{2.5cm}||M{2.5cm}|M{3.5cm}|M{1cm}|M{1cm}|M{1.5cm}| }
	    \hline
	     Name & No. of Params & Training Iterations & EM & F1 & Accuracy\\
	     \hline
	     \hline
	     Light QANet & 788,673 & 32,000 & 31.390 & 37.432 & 49.928 \\
	     \hline
	     Light QANet & 788,673 & 62,000 & 32.903 & 38.412 & 49.928 \\
	     \hline
	     EQuANt 1 & 996,196 & 40,000 & 32.881 & 38.356 & 49.914 \\
	     \hline
	     EQuANt 2 & 1,001,520 & 40,000 & 33.512 & 38.894 & 49.914 \\
	     \hline
	     EQuANt 3 & 927,970 & 62,000 & 56.843 & 60.980 & 69.114 \\
	     \hline
	     EQuANt 3 & 927,970 & 78,000 & 58.140 & 62.360 & 70.26 \\
	     \hline
	\end{tabular}
	\captionsetup{width=0.7\textwidth}
	\caption{SQuAD 2 dev set results.}
	\label{table:squad2results}
	\end{subfigure}
	
	\vspace{0.5em}
	\centering	
	\begin{subfigure}{2\columnwidth}
	\centering
	\begin{tabular}{ |c||c|c|c|c|c| }
	    \hline
	     Name & No. of Params & Training Iterations & EM & F1\\
	     \hline
	     \hline
	     Light QANet & 788,673 & 32,000 & 62.270 & 74.058 \\
	     \hline
	     Light QANet & 788,673 & 62,000 & 63.623 & 75.841 \\
	     \hline
	     EQuANt 3 & 927,970 & 62,000 & 69.29 & 78.80 \\
	     \hline
	\end{tabular}
	\captionsetup{width=0.7\textwidth}
	\caption{SQuAD 1.1 dev set results.}
	\label{table:squad1results}
	\end{subfigure}	
	
\end{figure*}

\section{Results \& Discussion}
\label{sec: results}

As mentioned in section \ref{subsec:methods_base_qa}, our first step was to train our light QANet on SQuAD 1.1 for $32,000$ iterations in order to generate a suitable initialisation which was used for the subsequent training of all other models. Evaluation of this trained model on SQuAD 1.1 yields the results shown in the first row of table \ref{table:squad1results}. The quoted number of training iterations for other models in tables \ref{table:squad1results} and \ref{table:squad2results} therefore includes these $32,000$ pre-training iterations. In order to observe how our lightweight model trained on SQuAD 1.1 without data augmentation compares to the original QANet without data augmentation, we trained for a further $30,000$ iterations on SQuAD 1.1, yielding the results shown in the second row of table \ref{table:squad1results}. These EM/F1 scores are 9.98/6.853 lower than the corresponding results for the full QANet architecture \cite{qanet}, implying that our choice to employ a lightweight architecture has a noticeable impact on performance.

We evaluated these trained light QANet models on the SQuAD 2 dev set, implicitly treating all questions as answerable. This led to the results shown in the first and second rows of table \ref{table:squad2results}, which act as baselines to compare our EQuANt results against. The accuracy column in table \ref{table:squad2results} contains the proportion of questions correctly identified as being answerable or unanswerable.

Having investigated the performance of light QANet on SQuAD 1.1 and 2, we moved on to train each of the EQuANt architectures described in section \ref{subsec: Architecture Design} on SQuAD 2. As can be seen in table \ref{table:squad2results}, EQuANt 1 and 2 did not perform well on SQuAD 2. In fact, their performance is identical. This is explained by both models learning to output a constant answerability probability of 0.69, independent of the query-context pair considered. Note that this probability matches the proportion of SQuAD2 training examples which are answerable, meaning that these models have been unable to extract the necessary features for accurately predicting answerability and have defaulted to the most basic frequentist approach of predicting the mean.

However, as shown in the final row of table \ref{table:squad2results}, EQuANt 3 is capable of both answerability prediction and question answering, significantly exceeding baseline performance on SQuAD 2.


As laid out in this \href{http://ruder.io/multi-task-learning-nlp/index.html}{blog post} by Sebastian Ruder, multi-task learning has recently been successfully applied to numerous NLP tasks. We therefore decided to measure the performance of EQuANt 3, trained on the two tasks of question answering and answerability prediction, at question answering alone by evaluating its EM and F1 scores on SQuAD 1 by providing EQuANt 3 with the ground truth answerability of true for each SQuAD1 question. As shown in the final row of table \ref{table:squad1results}, EQuANt 3 outperforms light QANet by 5.667/2.959 on F1/EM scores, suggesting that it indeed benefits from this multi-task approach.

\section{Conclusion}
\label{sec:Conclusion}
In this work, we have presented EQuANt, an MRC model which extends QANet to cope with unanswerable questions.
In sections \ref{sec:Background} and \ref{sec:Related Work}, we motivated our work and placed it in the wider
context of MRC and unanswerability. Following this, in section \ref{sec:Methods}, we presented our lightweight
QANet implementation and laid out in detail the 3 EQuANt architectures that were trained and whose performance was evaluated.
In section \ref{sec: experiment}, we investigated the context-query attention maps within our lightweight QANet,
allowing us to verify the quality of our learnt encodings and gain insight into why our initial architecture, EQuANt 1
did not predict answerability effectively. Finally, in section \ref{sec: results}, we presented our results and
discussed how the observed performance of EQuANt 3 on SQuAD 1.1 suggests that multi-task learning is a valuable
approach in the context of MRC.







\bibliographystyle{acl_natbib}
\bibliography{ourbib}

\begin{thebibliography}{15}
\expandafter\ifx\csname natexlab\endcsname\relax\def\natexlab#1{#1}\fi

\bibitem[{Devlin et~al.(2018)Devlin, Chang, Lee, and Toutanova}]{BERT}
Jacob Devlin, Ming{-}Wei Chang, Kenton Lee, and Kristina Toutanova. 2018.
\newblock \href {http://arxiv.org/abs/1810.04805} {{BERT:} pre-training of deep
  bidirectional transformers for language understanding}.
\newblock \emph{CoRR}, abs/1810.04805.

\bibitem[{Dhingra et~al.(2016)Dhingra, Liu, Cohen, and
  Salakhutdinov}]{dhingral16}
Bhuwan Dhingra, Hanxiao Liu, William~W. Cohen, and Ruslan Salakhutdinov. 2016.
\newblock \href {http://arxiv.org/abs/1606.01549} {Gated-attention readers for
  text comprehension}.
\newblock \emph{CoRR}, abs/1606.01549.

\bibitem[{Hill et~al.(2015)Hill, Bordes, Chopra, and Weston}]{hill16}
Felix Hill, Antoine Bordes, Sumit Chopra, and Jason Weston. 2015.
\newblock \href {http://arxiv.org/abs/1511.02301} {The goldilocks principle:
  Reading children's books with explicit memory representations}.
\newblock \emph{CoRR}, abs/1511.02301.

\bibitem[{Hu et~al.(2018)Hu, Wei, Peng, Huang, Yang, and Zhou}]{read_verify}
Minghao Hu, Furu Wei, Yuxing Peng, Zhen Huang, Nan Yang, and Ming Zhou. 2018.
\newblock \href {http://arxiv.org/abs/1808.05759} {Read + verify: Machine
  reading comprehension with unanswerable questions}.
\newblock \emph{CoRR}, abs/1808.05759.

\bibitem[{Levy et~al.(2017)Levy, Seo, Choi, and Zettlemoyer}]{Levy_et_al}
Omer Levy, Minjoon Seo, Eunsol Choi, and Luke Zettlemoyer. 2017.
\newblock \href {http://arxiv.org/abs/1706.04115} {Zero-shot relation
  extraction via reading comprehension}.
\newblock \emph{CoRR}, abs/1706.04115.

\bibitem[{Liu et~al.(2018)Liu, Li, Fang, Kim, Duh, and Gao}]{extended_SAN}
Xiaodong Liu, Wei Li, Yuwei Fang, Aerin Kim, Kevin Duh, and Jianfeng Gao. 2018.
\newblock \href {http://arxiv.org/abs/1809.09194} {Stochastic answer networks
  for squad 2.0}.
\newblock \emph{CoRR}, abs/1809.09194.

\bibitem[{Liu et~al.(2017)Liu, Shen, Duh, and Gao}]{SAN}
Xiaodong Liu, Yelong Shen, Kevin Duh, and Jianfeng Gao. 2017.
\newblock \href {http://arxiv.org/abs/1712.03556} {Stochastic answer networks
  for machine reading comprehension}.
\newblock \emph{CoRR}, abs/1712.03556.

\bibitem[{Pennington et~al.(2014)Pennington, Socher, and Manning}]{glove}
Jeffrey Pennington, Richard Socher, and Christopher~D. Manning. 2014.
\newblock \href {http://www.aclweb.org/anthology/D14-1162} {Glove: Global
  vectors for word representation}.
\newblock In \emph{Empirical Methods in Natural Language Processing (EMNLP)},
  pages 1532--1543.

\bibitem[{Rajpurkar et~al.(2018)Rajpurkar, Jia, and Liang}]{squad2}
Pranav Rajpurkar, Robin Jia, and Percy Liang. 2018.
\newblock \href {http://arxiv.org/abs/1806.03822} {Know what you don't know:
  Unanswerable questions for squad}.
\newblock \emph{CoRR}, abs/1806.03822.

\bibitem[{Rajpurkar et~al.(2016)Rajpurkar, Zhang, Lopyrev, and Liang}]{squad1}
Pranav Rajpurkar, Jian Zhang, Konstantin Lopyrev, and Percy Liang. 2016.
\newblock \href {http://arxiv.org/abs/1606.05250} {Squad: 100, 000+ questions
  for machine comprehension of text}.
\newblock \emph{CoRR}, abs/1606.05250.

\bibitem[{Seo et~al.(2016)Seo, Kembhavi, Farhadi, and Hajishirzi}]{BiDAF}
Min~Joon Seo, Aniruddha Kembhavi, Ali Farhadi, and Hannaneh Hajishirzi. 2016.
\newblock \href {http://arxiv.org/abs/1611.01603} {Bidirectional attention flow
  for machine comprehension}.
\newblock \emph{CoRR}, abs/1611.01603.

\bibitem[{Sun et~al.(2018)Sun, Li, Qiu, and Liu}]{unet}
Fu~Sun, Linyang Li, Xipeng Qiu, and Yang Liu. 2018.
\newblock \href {http://arxiv.org/abs/1810.06638} {U-net: Machine reading
  comprehension with unanswerable questions}.
\newblock \emph{CoRR}, abs/1810.06638.

\bibitem[{Vaswani et~al.(2017)Vaswani, Shazeer, Parmar, Uszkoreit, Jones,
  Gomez, Kaiser, and Polosukhin}]{need_paper}
Ashish Vaswani, Noam Shazeer, Niki Parmar, Jakob Uszkoreit, Llion Jones,
  Aidan~N Gomez, \L~ukasz Kaiser, and Illia Polosukhin. 2017.
\newblock \href
  {http://papers.nips.cc/paper/7181-attention-is-all-you-need.pdf} {Attention
  is all you need}.
\newblock In I.~Guyon, U.~V. Luxburg, S.~Bengio, H.~Wallach, R.~Fergus,
  S.~Vishwanathan, and R.~Garnett, editors, \emph{Advances in Neural
  Information Processing Systems 30}, pages 5998--6008. Curran Associates, Inc.

\bibitem[{Wang et~al.(2017)Wang, Yang, Wei, Chang, and Zhou}]{rnet}
Wenhui Wang, Nan Yang, Furu Wei, Baobao Chang, and Ming Zhou. 2017.
\newblock Gated self-matching networks for reading comprehension and question
  answering.
\newblock In \emph{Proceedings of the 55th Annual Meeting of the Association
  for Computational Linguistics (Volume 1: Long Papers)}, volume~1, pages
  189--198.

\bibitem[{Yu et~al.(2018)Yu, Dohan, Luong, Zhao, Chen, Norouzi, and Le}]{qanet}
Adams~Wei Yu, David Dohan, Minh{-}Thang Luong, Rui Zhao, Kai Chen, Mohammad
  Norouzi, and Quoc~V. Le. 2018.
\newblock \href {http://arxiv.org/abs/1804.09541} {Qanet: Combining local
  convolution with global self-attention for reading comprehension}.
\newblock \emph{CoRR}, abs/1804.09541.

\end{thebibliography}


\end{document}